\documentclass[letterpaper]{article}
\usepackage{flairs}
\usepackage{times}
\usepackage{helvet}
\usepackage{courier}
\usepackage{graphicx}
\usepackage{amsmath}
\usepackage{amsfonts}
\begin{document}
%
\title{3M: Multi-style image caption generation using Multi-modality features under Multi-UPDOWN model}
\author{Chengxi Li and Brent Harrison\\
University of Kentucky\\
Lexington, KY 40506\\
}
\maketitle
\begin{abstract}
\begin{quote}
 In this paper, we build a multi-style generative model for stylish image captioning which uses multi-modality image features, ResNeXt features and text features generated by DenseCap. We propose  the 3M model, a Multi-UPDOWN caption model that encodes multi-modality features and decode them to captions. We demonstrate the effectiveness of our model on generating human-like captions by examining its performance on two datasets, the PERSONALITY-CAPTIONS dataset and the FlickrStyle10K dataset. We compare against a variety of state-of-the-art baselines on various automatic NLP metrics such as BLEU, ROUGE-L, CIDEr, SPICE, etc \footnote{code will be available at https://github.com/***}. A qualitative study has also been done to verify our 3M model can be used for generating different stylized captions.
\end{quote}
\end{abstract}
\section{Introduction}
Factual image captioning is one of the fundamental tasks in deep learning. 
The issue with factual captions is that language generated is often stilted, and not necessarily representative of human communication. 
While classic image captioning approaches show deep understanding of image composition and language construction, it often lacks elements that make communication distinctly human. 
To address this issue, some researchers have tried to add personality to image captioning in order to generate stylish captions.
In general, stylish captioning systems are divided into two categories based on how they are trained: 
single style \cite{gan2017stylenet,zhang2018style} and multi-style \cite{shuster2019engaging,guo2019mscap,zhao2020memcap}. 
Single-style training involves training one model for each personality, whereas multi-style techniques learn to generate captions in many different styles using one model. 

Shuster \textit{et al.} built a multi-style module by converting each personality to a multi-dimensional vector. 
Their generative model struggled to generate captions that accurately captured the given image context. 
This is likely because a multi-style captioners require greater knowledge about the input image when compared to single-style captioners. 
To address the inherent limitations of past multi-style captioning approaches, we propose the use of multi-modality image features to improve the quality of multi-style image captioning. 
We believe that multi-modality features, specifically image features combined with features derived from text describing said image, will help the model better ground image features into text. 

To effectively generate stylish captions, a model needs to incorporate elements of the local context of image regions and the global context of the image itself. 
To capture local context, our model will 
make use of region-based caption features generated by the DenseCap network~\cite{densecap}. 
To complement dense caption features, we will also use ResNext features describing the global input image. 
To combine these features, we introduce a Multi-UPDOWN structure model where each UPDOWN structure is used to select the best feature from its own modality. 
These selections are then fused to generate the caption. 

To evaluate the performance of our multi-style captioning model, we examine its performance on different stylish image captioning datasets. 
We evaluate its performance using various NLP metrics and compare against several state-of-the-art baselines. 
We perform an ablation study in which we examine how each part of our model contributes to the overall expressiveness and diversity of our generated captions. 
We also perform a qualitative evaluation in which we examine how well the captions generated by our model capture image and style context.






\begin{figure*}[t]
\center
\includegraphics[scale=0.49]{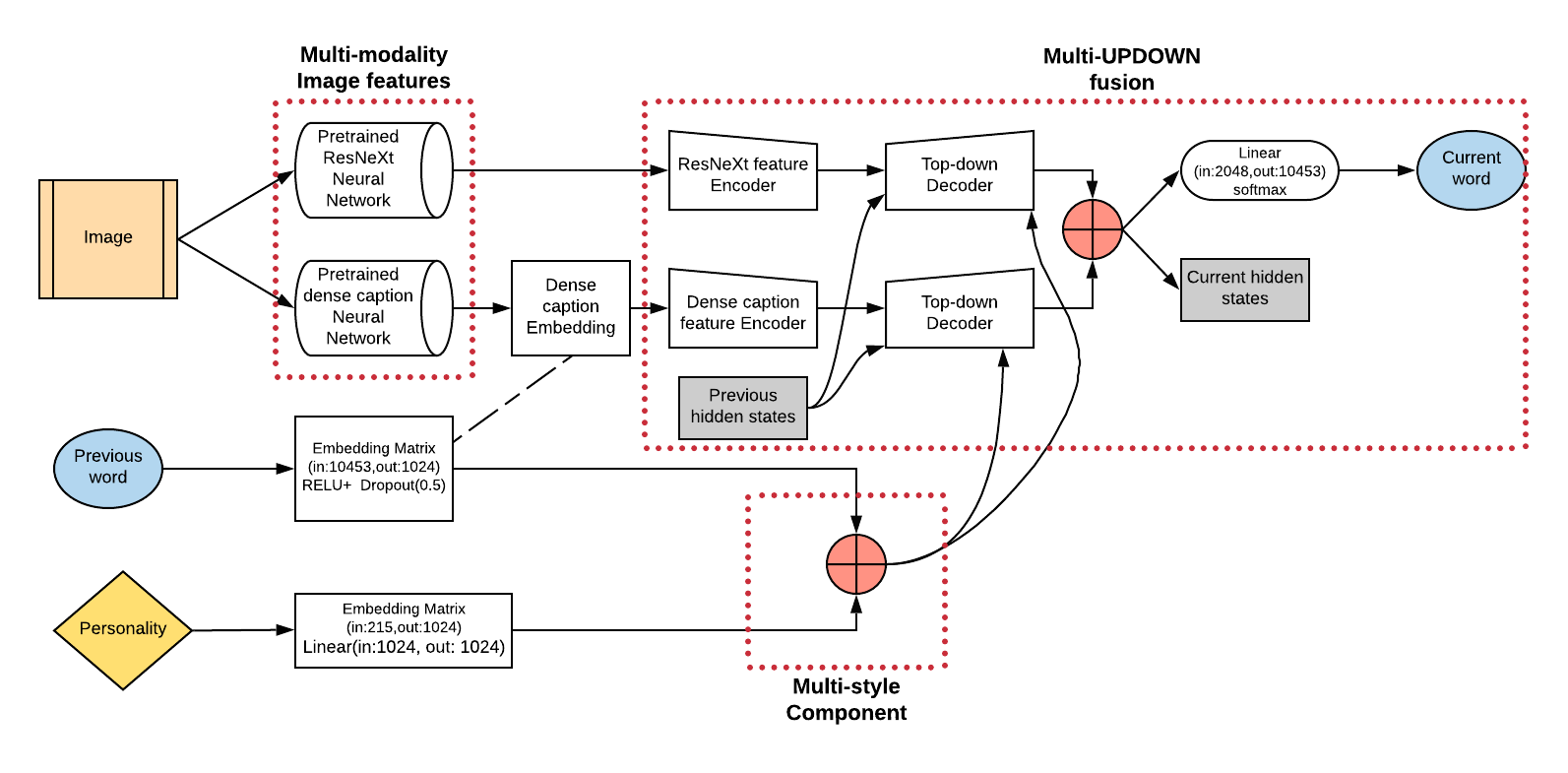}
\caption{Architecture for Multi-style image caption generation using Multi-modality features under Multi-UPDOWN model}
\label{fig:architecture}
\end{figure*}


\begin{figure}[t]
\centering
\includegraphics[width=\columnwidth]{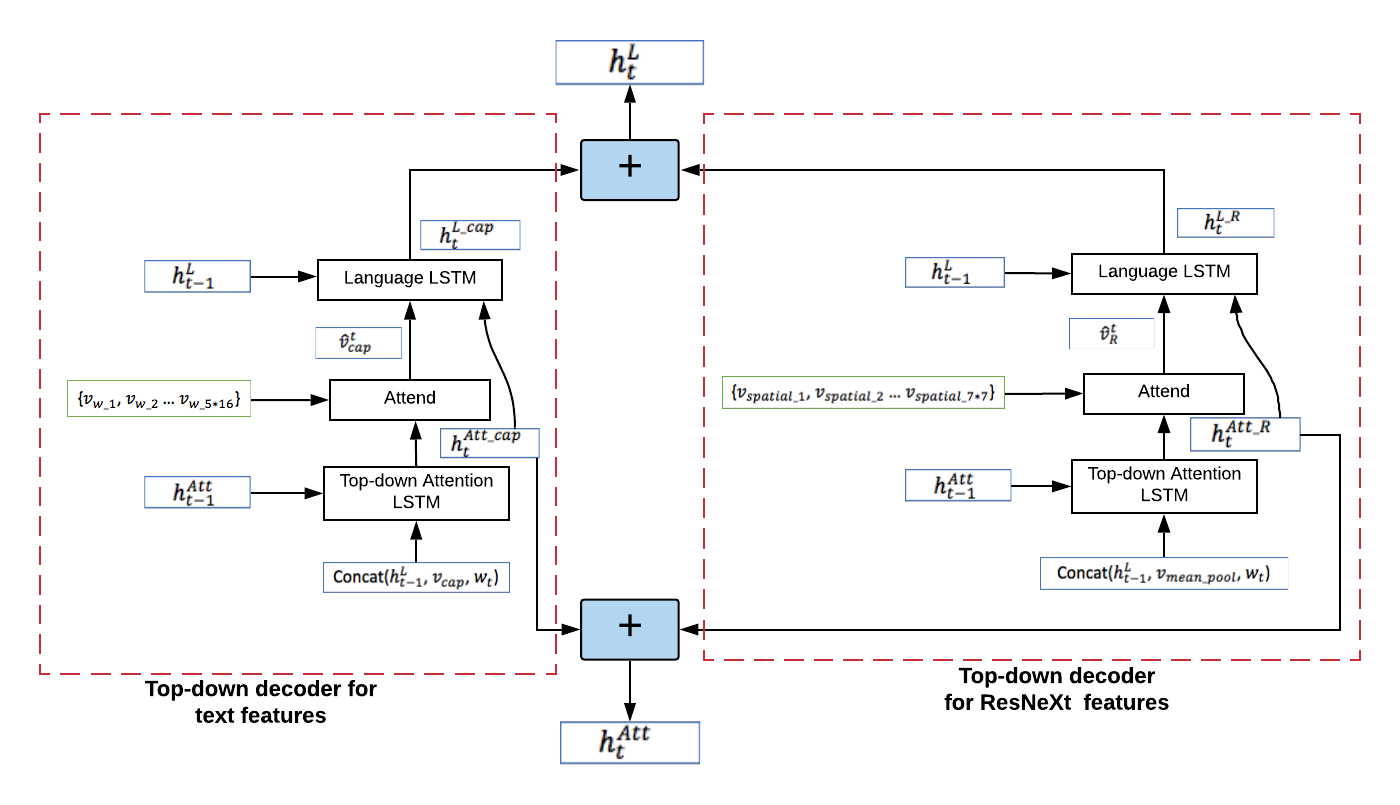}
\caption{Two Decoders Fusion Details}
\label{fig:two_updown}
\end{figure}
\section{Related Work}
Captions in FlickrStyle10K are created to have either a Humorous or Romantic linguistic style ~\cite{gan2017stylenet} while captions in PERSONALITY-CAPTIONS are created to be engaging and have a conversational style \cite{shuster2019engaging}.
With FlickrStyle10K, researchers have built single-style captioners \cite{gan2017stylenet,chen2018factual} where they make use of both factual captions and stylized captions for training. 
Later researchers explored training multi-style networks \cite{guo2019mscap,zhao2020memcap} that can generate multiple types of stylish outputs using a single model. 

Shuster \textit{et al.} released the PERSONALITY-CAPTIONS containing 215 personalities in 2019 for building engaging caption generations models. In their work, Shuster \textit{et al.} built an image caption retrieval model and also explored the multi-style generative caption models along with various image encoding strategies ~\cite{he2016deep,xie2017aggregated}
using several state-of-the-art image captioning models~\cite{xu2015show,anderson2018bottom}. 
They applied a supervised learning model plus reinforcement learning training strategy~\cite{rennie2017self} using CIDEr score~\cite{vedantam2015cider} as reward. 
We extend the best performing supervised model presented in Shuster \textit{et al.}'s work to build a multi-style model which incorporates multi-modality image features.


\section{Method}
The primary contribution of this paper is an architecture that utilizes multi-modality fusion for performing multi-style image captioning. 
This architecture specifically utilizes the soft fusion of two parallel encoder-decoder blocks, with each block containing an UPDOWN-like attention module. 
Our overall architecture for one step generation can be seen in Figure~\ref{fig:architecture}.
As seen in the figure, our multi-UPDOWN fusion blocks synthesize the information from multi-modality image features, multi-style components (previous word, personality) and previous hidden states to predict current word and hidden states at each time step.

In this architecture, we utilize two features from pre-trained networks: ResNeXt \shortcite{xie2017aggregated} visual features and text features describing the image itself \shortcite{densecap}.
These features allow the learner to better ground the image features into natural language.
\subsection{Multi-style Component}
As in the Figure \ref{fig:architecture}, the desired style of the output caption is given as an input to our system using a one-hot vector. 
We then use an embedding matrix $W_{p\_embed}$ and a linear layer to encode each style into a fixed-size vector, we call it style vector $p$. 
For each word in our target stylized caption, we use another embedding matrix $W_{embed}$ to embed each word. We will use $W_{embed}$ to embed the dense captions too. 
This enables us to better connect image features to natural language.
To better enable our network to generate words according to the given style, we concatenate each embedded word vector with the $p$ to create a stylized word vector, $\pmb{w}_{t}$.

\subsection{Multi-modality Image Features}
Our architecture relies on two sets of bottom-up features extracted using pre-trained networks: ResNeXt features and dense caption features. 
Specifically, we extract mean-pooled image features and spatial features from the ResNeXt network \cite{shuster2019engaging} and 5 dense captions from each image with a dense caption network \cite{densecap}. Each word in the dense captions is embedded using $W_{embed}$.
By collecting both visual and text features, we provide our architecture with a more complete understanding of the full context of the image. 

\subsection{Multi-UPDOWN fusion Model}
Our fusion model is composed of two individual encoders, the ResNext feature encoder and the dense caption encoder.
Our model also uses a fused Top-down fashion decoder, which used to decode captions from encoded image features.
\subsubsection{ResNeXt Feature Encoder and Dense Caption Encoder}
We encode the ResNeXt mean-pooled image features and spatial features using a linear layer, dropout layer and activation layer and get mean-pooled feature vector $\pmb{v}_{mean\_pool}$ and spatial feature vector $\pmb{v}_{spatial\_1}$, $\pmb{v}_{spatial\_2}$, 
..., $\pmb{v}_{spatial\_7*7}$. These are used as input features for the decoding process showed in the right branch of Figure \ref{fig:two_updown}. Then, we encode each embedded caption vector $Cap_{i}, i\in \{1,2,3,4,5\}$ using the Dense Caption Encoder, which is an LSTM network \cite{hochreiter1997long} shown below where $\pmb{w}^{dp}_{t,i}$ denotes a word vector in  $Cap_{i}$ at time t.
\begin{equation}\label{denseEncoder}
\footnotesize
\pmb{h}_{t,i}^{dp}, \pmb{c}_{t,i}^{dp}=LSTM(\pmb{w}^{dp}_{t,i},(\pmb{h}_{t-1,i}^{dp},\pmb{c}_{t-1,i}^{dp}))
\end{equation}
We concatenate all 5 hidden states $\pmb{h}_{i}^{dp}$ 
into one vector $\pmb{v}_{cap}$, which we call the \textit{caption vector}. 
To apply attention on specific words during the decoding procedure, we keep all word states $\pmb{c}_{t,i}^{dp}$  from the LSTM encoding process denoted as $\pmb{v}_{w_{1}},\pmb{v}_{w_{2}}...\pmb{v}_{w_{L}}$ where 5 captions contain total $L$ words.
\begin{table*}[t]
\footnotesize
 \centering
 \tabcolsep=0.10cm
 \begin{tabular}{llllllllll}
 \hline
 \textbf {Method}&  \textbf{Caption Model}&\textbf{Training Method}& \textbf{Text Features} &\textbf{ResNeXt}& \textbf{ BLEU1} & \textbf{ BLEU4}& {ROUGE-L}& \textbf{CIDEr}& \textbf{SPICE}\\
 \hline
 UPDOWN \shortcite{shuster2019engaging} & UPDOWN  & Supervised+Reinforcement & No & Yes & \bf 44.0 & 8.0 &27.4& 16.5 & \bf 5.2 \\
   UPDOWN \shortcite{shuster2019engaging} & UPDOWN  & Supervised & No & Yes & 40.5 & 6.9 &26.2& 16.2 & 4.0 \\
   \bf 3M & \bf Multi-UPDOWN  & Supervised & Yes & Yes &  43.0 & \bf 8.0&  \bf 27.6 & \bf 18.6 & 4.8\\
   \hline
\end{tabular}
 \caption{Performance of Generative Models on PERSONALITY-CAPTIONS Dataset. Note: Results of \cite{shuster2019engaging} under supervised learning are from re-training due to performance on supervised method not reported in \cite{shuster2019engaging} and some data of original dataset not available. We also listed original result of \cite{shuster2019engaging} which is under supervised and reinforcement learning for reference.}
 \label{tab:PPerformance}
\end{table*}
\begin{table*}[t]
 \centering
 \footnotesize
 \tabcolsep=0.10cm
 \begin{tabular}{llllllllll}
   \hline
   {\bf Caption Model}& {\bf Personality}& {\bf Text Features} & {\bf ResNeXt}& {\bf BLEU1} & {\bf BLEU4}& {\bf ROUGE-L}& {\bf CIDEr}& {\bf SPICE} &{\bf Unique words(\#)}\\
   \hline
     Multi-UPDOWN  & \bf No & Yes & Yes &  34.0 &  3.5&   22.3 &  11.1 & 3.6 & 257\\
   UPDOWN & Yes& \bf No & Yes & 42.4 & 7.5 &26.7& 17.9 & 4.4 & \bf 1558 \\
   UPDOWN & Yes & Yes & \bf No & \bf 43.2 & \bf 8.1&  \bf 27.6 &  18.0 &  4.6 & 1048\\
   Multi-UPDOWN  & Yes & Yes & Yes &  43.0 &  8.0&   \bf 27.6 & \bf 18.6 & \bf 4.8 & 1378 \\
 \hline
\end{tabular}
 \caption{Results of Ablation Studies on PERSONALITY-CAPTIONS Dataset}
 \label{tab:Ablation}
\end{table*}

\begin{table*}[t]
\centering
\footnotesize
  \tabcolsep=0.11cm
 \begin{tabular}{lllllllll}
   \hline
   {\bf Method}& {\bf style} &{\bf training method}&  {\bf BLEU1} & {\bf BLEU3}& {\bf Meteor}& {\bf CIDEr}& {\bf ppl}& {\bf cls}\\
   \hline
   SF-LSTM \shortcite{chen2018factual} & romantic &single-style& \bf 27.8 & \bf 8.2 & \bf 11.2 \bf & \bf 37.5&-&-\\
   SF-LSTM \shortcite{chen2018factual} & humorous &single-style& \bf 27.4 & \bf 8.5 & \bf 11.0 \bf & \bf 39.5&-&-\\   
   StyleNet \shortcite{gan2017stylenet} & romantic & single-style & 13.3 & 1.5 &4.5& 7.2&52.9&37.8\\
   StyleNet \shortcite{gan2017stylenet} & humorous & single-style & 13.4 & 0.9 &4.3& 11.3&48.1 &41.9\\
   MsCap \shortcite{guo2019mscap} & romantic & multi-style & 17.0 & 2.0 &5.4& 10.1& 20.4 &88.7\\
   MsCap \shortcite{guo2019mscap} & humorous & multi-style & 16.3 & 1.9 &5.3& 15.2& 22.7 &91.3\\
   MemCap \shortcite{zhao2020memcap} & romantic & multi-style & 19.7 & 4.0 &7.7& 19.7& 19.7& 91.7\\ 
   MemCap \shortcite{zhao2020memcap} & humorous & multi-style & 19.8 & 4.0 &7.2& 18.5&17.0& \bf 97.1\\ 
   \bf 3M & romantic & multi-style & 25.6 &  6.7 &   10.1 &  29.3 & \bf 8.33 & \bf 92.8\\
   \bf 3M & humorous & multi-style & 25.5 &  6.7 &  10.0 & 28.4& \bf 7.29 & 95.3\\
 \hline
\end{tabular}
 \caption{Performance of Generative Models on FlickrStyle10K Dataset. Note: Due to only 7K out of 10K dataset publicly available, all the result are reported based on 7k data. All results except 3M are referred from paper \cite{zhao2020memcap}.}
 \label{tab:FPerformance}
\end{table*}
\begin{figure*}[!ht]
 \centering
 \includegraphics[scale=0.3]{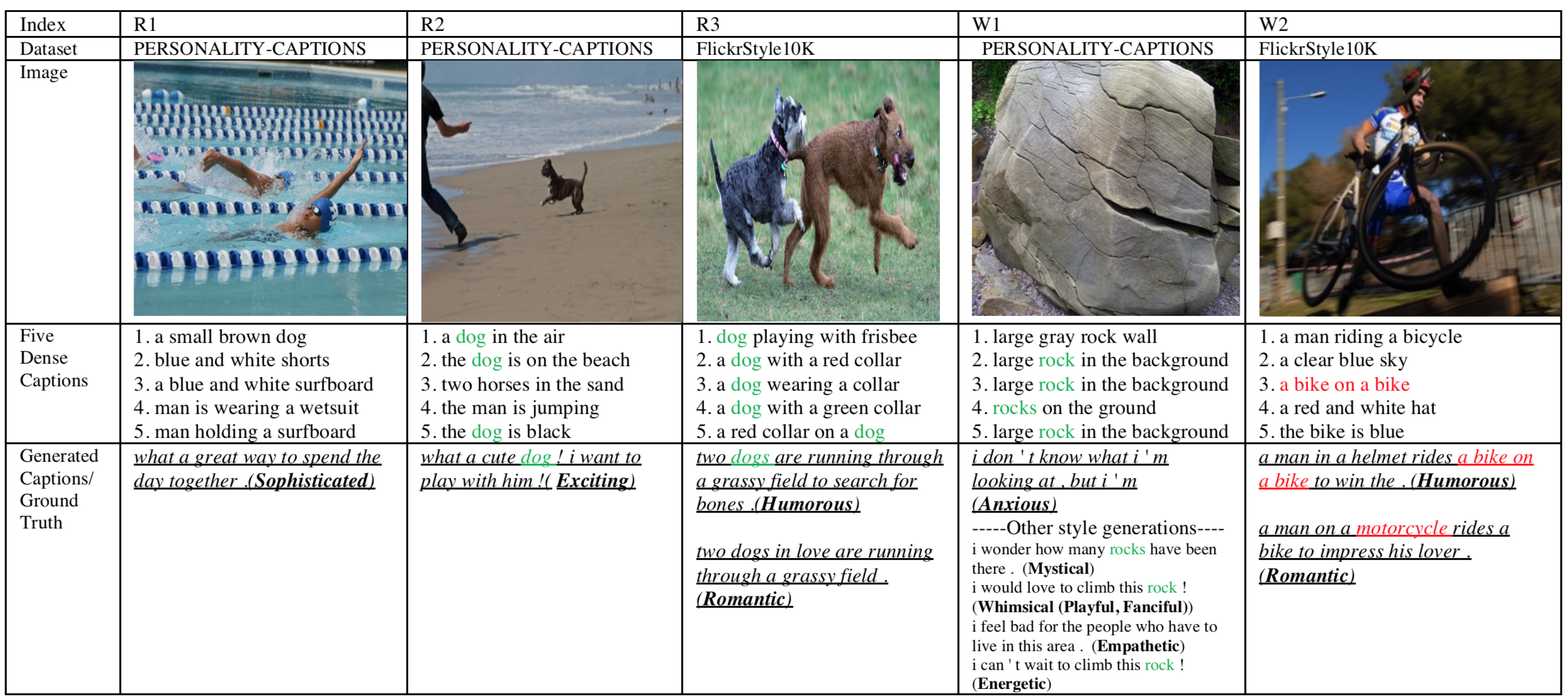}
 \caption{R1-R3: Generated Captions samples using 3M trained on PERSONALITY-CAPTIONS and FlickrStyle10K (underscored).
W1-W2: Imperfect Captions generations samples using 3M trained on PERSONALITY-CAPTIONS and FlickrStyle10K (underscored) along with generations from the same image and other personalities, personality are listed in parenthesis, ground truth has the same personality as the underscored generations}
 \label{fig:rightwrong}
\end{figure*}
\subsubsection{Top-down Decoder Fusion}
As we show in Figure \ref{fig:two_updown}, we apply the Top-down decoder model on encoded visual features and text features. 
The left branch is the Top-down decoder for our text features generated by the dense caption network and the right branch is the Top-down decoder for the ResNeXt features.
At each time step, the Top-down decoder for text features generates a caption attention vector $\pmb{h}_{t}^{Att\_cap}$ by taking in the previous attention vector hidden states $\pmb{h}_{t-1}^{Att}$ as well as the concatenation of previous language model hidden states $\pmb{h}_{t-1}^{L}$, the caption vector $\pmb{v}_{cap}$ and the previous stylized word vector $\pmb{w}_{t}$ as input.
\begin{equation}
\label{TopDownAttLSTM}
\footnotesize
\pmb{h}_{t}^{Att\_cap}=TopDownAttLSTM([\pmb{h}_{t-1}^{L},\pmb{v}_{cap},\pmb{w}_{t}],\pmb{h}_{t-1}^{Att})
\end{equation}
To calculate the \textit{attended caption feature vector} We use a process inspired by \cite{anderson2018bottom}.
We use vectors $\pmb{v}_{w_{1}},\pmb{v}_{w_{2}}...\pmb{v}_{w_{L}}$ and the caption attention vector $\pmb{h}_{t}^{Att\_cap}$ in the below equations: 
\begin{align}\label{AttendAtt}
\footnotesize
& a_{i,t}= \pmb{w}_{a}^{T}tanh(W_{va}\pmb{v}_{w_{i}} + W_{ha}\pmb{h}_{t}^{Att\_cap}
) \\
& \pmb{\alpha}_{t} = softmax (\pmb{a}_{t}) \\
& \widehat{\pmb{v}}_{cap}^{t} =\sum_{i=1}^{K}{\pmb{\alpha}_{i}^{t}\pmb{v}_{w_{i}}}
\end{align}
where $W_{va} \in \mathbb{R}^{H\times V}, W_{ha}\in  \mathbb{R}^{H\times M}$ and $\pmb{w}_{a} \in \mathbb{R}^{H}$ are learned parameters.
This attention vector $\widehat{\pmb{v}}_{cap}^{t}$ is used as the input to the language LSTM layer where the initial state is the previous hidden state from the language model, $\pmb{h}_{t-1}^{L}$. 
This language LSTM then outputs the current language model hidden states $\pmb{h}_{t}^{L\_cap}$ for our text features as below:
\begin{equation}\label{LangLSTM}
\footnotesize
\pmb{h}_{t}^{L\_cap}=LanguageLSTM([\widehat{\pmb{v}}_{cap}^{t},\pmb{h}_{t}^{Att\_cap}],\pmb{h}_{t-1}^{L})
\end{equation}
We calculate the ResNeXt attention vector $\pmb{h}_{t}^{Att\_R}$, and current language model hidden states from ResNeXt features $\pmb{h}_{t}^{L\_R}$, using a similar process with a separate network (shown in Figure~\ref{fig:two_updown} right branch).
We generate the final language hidden states of the current step $\pmb{h}_{t}^{L}$ by fusing $\pmb{h}_{t}^{L\_cap}$, $\pmb{h}_{t}^{L\_R}$ as below:
\begin{equation}\label{eqfuse1}
\footnotesize
\pmb{h}_{t}^{L}= \pmb{h}_{t}^{L\_cap}+ \pmb{h}_{t}^{L\_R}
\end{equation}
We generate the final attention hidden states of the current step $\pmb{h}_{t}^{Att}$ by fusing  $\pmb{h}_{t}^{Att\_cap}$, $\pmb{h}_{t}^{Att\_R}$ as below:
\begin{equation}\label{eqfuse2}
\footnotesize
\pmb{h}_{t}^{Att}= \pmb{h}_{t}^{Att\_cap}+ \pmb{h}_{t}^{Att\_R}
\end{equation}
We get the final language output as below:
\begin{equation}\label{eqfuse3}
\footnotesize
\pmb{h}_{t}^{output}=Dropout(\pmb{h}_{t}^{L\_cap})+ Dropout(\pmb{h}_{t}^{L\_R})
\end{equation}
Then we apply a linear layer to project the final language output $\pmb{h}_{t}^{output}$ to the vocabulary space and use a log softmax layer to convert it to a log probability distribution.


\section{Experimental Methodology}
To demonstrate the effectiveness of our model on stylish image captioning, we use the PERSONALITY-CAPTIONS dataset, which contains 215 distinct personalities. 
To prove our model is expandable to linguistic stylized captions, we train our model using the FlickrStyle10K dataset \cite{gan2017stylenet} which contains humorous and romantic personalities. 
We compare our results with the state-of-the-art work on the same datasets based on their automatic evaluation metrics. Ablation studies are also done to justify the contributions of each component of our method. We also perform a qualitative examination of the outputs of our model. 
\subsection{Dataset Details}

The ground truth captions in PERSONALITY-CAPTIONS \cite{shuster2019engaging} are created to be engaging and have a human-like style.
Each data entry in this dataset is represented as a triple containing an image, personality trait, and caption. 
In total, 241,858 captions are included in this dataset. 
In this work, we do not use the full PERSONALITY-CAPTIONS dataset due to accessibility of some examples. 
In total, our reduced dataset contains 186698 examples in the training set, 4993 examples in the validation set, and 9981 examples in the test set. The total vocabulary size of PERSONALITY-CAPTIONS after replacing infrequent tokens with 'UNK' is 10453.

The FlickrStyle10K dataset captions focus on linguistic style.
Since only 7000 images are publicly available, we evaluate using a similar process to the one outlined in \cite{guo2019mscap,zhao2020memcap}.
First we randomly select 6,000 images as the training data and use the remaining 1000 images as testing data. We further split 10\% data from training data as validation data. 
Total vocabulary size of FlickrStyle10K is 8889.

\subsection{Training and Inference}
In the training, we use entropy as loss function and Adam optimization with initial learning rate of 5e-4. 
The learning rate decays every 5 epochs. 
In total, we train 30 epochs when using the PERSONALITY-CAPTIONS dataset~\cite{shuster2019engaging} with a batch size of 128 and evaluate the model every 3000 iterations. We train for 100 epochs when using the FlickrStyle10K dataset \cite{gan2017stylenet} with batch size 128 and evaluate model every 100 iterations. 

During inference, we generate captions using beam search with beam size 5. During this process, we impose a penalty to discourage the network from repeating words, from ending on words such as an, the, at, etc and from generating special tokens, like 'UNK'.

\subsection{Quantitative Analysis}
Our quantitative analysis is meant to show that our 3M model can outperform several state-of-the-art baselines in terms of a set of automated NLP metrics. 
In addition, we run an ablation study to validate the need for each part of the 3M model. 
\subsubsection{Baselines and Evaluation Metrics}
To test if our 3M model can be used to generate human-like captions, we train it using the above settings on the PERSONALITY-CAPTIONS dataset. 
We compare against the model introduced previously by Shuster \textit{et al.}~\cite{shuster2019engaging}. 
Since we use a subset of the original PERSONALITY-CAPTIONS dataset, we retrain the method outlined by Shuster \textit{et al.} using similar settings. 
We compare the performance of our 3M model against their model using  using BLEU~\cite{papineni2002bleu}, ROUGE-L~\cite{lin-2004-rouge}, CIDEr~\cite{vedantam2015cider}, and SPICE~\cite{anderson2016spice}. The comparison results are listed in Table \ref{tab:PPerformance}.

To evaluate the extensibility of our model, we also applied our method on the FlickrStyle10K dataset. 
This is meant to evaluate how well our method can generate captions that capture linguistic style. 
We compare against the following state-of-the-art baselines:
\begin{itemize}
  \item StyleNet \shortcite{gan2017stylenet}, a single style model trained with paired factual sentences and unpaired stylized captions.
  \item SF-LSTM \shortcite{chen2018factual}, a single style model trained with paired stylized caption and paired factual captions.
  \item MsCap \shortcite{guo2019mscap}, a multi-style model trained with paired factual sentences and unpaired stylized captions.
  \item MemCap \shortcite{zhao2020memcap}, a multi-style model trained with paired factual sentences and unpaired stylized captions.
\end{itemize}
Following \cite{zhao2020memcap}, on FlickrStyle10K, we trained a logistic regression classifier for style classification and a pretrained language model using SRILM toolkit \cite{stolcke2002srilm} to measure perplexity. We report BLEU, Meteor \cite{banerjee-lavie-2005-meteor}, CIDEr, the style classification accuracy (cls) and the average perplexity (ppl) for comparison and results are showed in Table \ref{tab:FPerformance}.

\subsubsection{Ablation Study}
Additionally, to evaluate the benefits of each component of our model, we perform an ablation study using the PERSONALITY-CAPTIONS dataset. 
We compare the full 3M model against the following variations: no personality features, no text features, and no ResNeXt features. 
BLEU, ROUGE-L, CIDEr, and SPICE are reported in Table \ref{tab:Ablation} for evaluating the relevance between image and generations.
we also report the number of unique words used across all generated captions per model in Table \ref{tab:Ablation} to show the expressiveness of each generative model. 

\subsection{Qualitative Analysis}
Specifically, we seek to explain that our model is capable of generating captions that match the given style as well as the image context.
We first list the given image and five given dense captions, sample generations along with personality in the parenthesis,  in Figure \ref{fig:rightwrong} as R1-R3.
We discuss the whether caption generations matching the context in three aspects: 1. whether the multi-style component working for connecting caption generations with given personality; 2. whether valid text features could help for generations to match the image; 3. whether ResNext feature could help make reasonable generations when the given text features fails to connect with the image. 
To give a more complete view of the text that our model can generate, we also list the imperfect sample generations underlined in Figure \ref{fig:rightwrong} as W1-W2. 



\section{Results and Discussion}
In this section, we will outline the results of our experiments and illustrate them in both quantitative and qualitative ways.
\subsection{Quantitative Analysis}
\subsubsection{Comparison with baselines}
As seen in Table \ref{tab:PPerformance}, our 3M model outperforms UPDOWN models under the same training method across all the NLP metrics we used for evaluation.
We also achieve better results on ROUGE-L, CIDEr compared with Shuster's model trained under reinforcement learning. 
This provides evidence that our approach is effective at multi-style caption generation. 

We also show that our 3M model does well on linguistic style captioning even though it was not designed for that task. 
As Table \ref{tab:FPerformance} shows, our 3M model significantly outperforms two other multi-style models, MsCap \cite{guo2019mscap} and MemCap \cite{zhao2020memcap} on BLEU, CIDEr, Meteor, and ppl on the FlickrStyle10K dataset. 
Note that our 3M model also achieved high cls values, which show how well our captions capture the given style. 

We also achieve comparable performance to the SF-LSTM model across the automated metrics we examined. 
Given that the SF-LSTM model is designed for a single-style generation task, whereas our 3M model was designed for multi-style generation, we feel that this shows how robust our model is. 

\subsubsection{Ablation Study}
From Table \ref{tab:Ablation}, we can see if our model is trained without the multi-style component, the performance of all the nlp metrics drops, proving how critical this component is. 
Examining the results obtained from a model using only text features against a model that only had access to ResNeXt features shows that using only text features limits the overall expressiveness of generated captions as shown by the low number of unique words generated.

Our full model has achieved the highest ROUGE-L, CIDEr and SPICE score and improves expressiveness compared with model with only text features and improves the relevancy compared to a model with only Resnext features. 
\subsection{Qualitative Analysis}
For our qualitative analysis, we will discuss the quality of the trained 3M models across two datasets assessing whether our model is capable of generating captions that match the given style and image context, and assessing whether our model can assist in finding reasons for imperfect captions. 

From all generations in Figure \ref{fig:rightwrong}, we can see our 3M model is able to generate captions matching the given personality, which certify that our multi-style component is able to help direct the generations in the desired personality tone. From R2-R3 we can see that when there is a valid text feature available, the 3M model could make use of them. The generation in R1 is expressed in a more conservative and global way since text features cannot provide correct information, which necessitates the use of ResNext features. 

One of the advantages of the 3M model is that it can easily generate multiple captions with different styles. This can enable us to better contextualize incomplete or erroneous captions. In W1 of Figure \ref{fig:rightwrong}, for example, the generation appear incomplete for the ``Anxious'' personality. Looking at captions for other personalities, we see that our model can correctly identify image context. This leads us to believe that we simply set the caption length too low for the ``anxious'' example. In W2, our model generates the incorrect phrases ``a bike on a bike.'' By examining the text features used for generation, we can see that this was likely caused by our input text, and not the model itself. 

\section{Conclusion}
In this paper we introduce the 3M model, which is a multi-style image captioner which integrates multi-modal features and a multi-UPDOWN encoder-decoder model. 
We demonstrate the effectiveness of our 3M model by comparing against state-of-the-art work using automatic evaluation methods. Ablation studies have also be done to evaluate the contributions of each component of our 3M model. And we certify that our 3M model could generate more expressive and diverse generations without losing the connection with context. The qualitative study helps understand how well our 3M performs and shows how our model can also explain the imperfectness of generations.

\bibliographystyle{flairs}
\bibliography{3m-paper.bib}

\end{document}